# Appropriateness is all you need!

### General-purpose chatbots and what they may and may not say

**Hendrik Kempt**[1], **Alon Lavie**[2,3], **Saskia K. Nagel**[1]

[1]RWTH Aachen University, Applied Ethics Group

[2]Carnegie Mellon University, Language Technologies Institute

[3]Unbabel Inc.

`{hendrik.kempt, saskia.nagel}@humtec.rwth-aachen.de`

**Abstract**

The strive to make AI applications "safe" has led to the development of safety-measures as the main or even sole normative requirement of their permissible use. Similar can be attested to the latest version of chatbots, such as chatGPT. In this view, if they are "safe", they are supposed to be permissible to deploy. This approach, which we call "safety-normativity", is rather limited in solving the emerging issues that chatGPT and other chatbots have caused thus far. In answering this limitation, in this paper we argue for limiting chatbots in the range of topics they can chat about according to the normative concept of *appropriateness*. We argue that rather than looking for the "safety" in a chatbot's utterances to determine what they may and may not say, we ought to assess those utterances according to three forms of appropriateness: technical-discursive, social, and moral. We then spell out what requirements for chatbots follow from these forms of appropriateness to avoid the limits of previous accounts: positionality, acceptability, and value alignment (PAVA). With these in mind, we may be able to determine what a chatbot may or may not say. Lastly, one initial suggestion is to use challenge sets, specifically designed for appropriateness, as a validation method.

**1. Introduction**

General-purpose and open-domain chatbots have entered the public domain, to varying degrees of economic success and public acclaim. OpenAI's recent public release of ChatGPT (Schulman et al. 2022), a chatbot designed and built on top of the latest generation of large language models (LLMs), has stunned the public in its capabilities, and generated a deluge of public discourse. Two intuitively plausible demands now stand against each other: on the one hand, chatbots ought not to say just anything, as some speech can cause great harm. On the other hand, for general purpose use and to be as helpful as possible, they ought to remain open-domain chatbots. Walking the line between ensuring as much freedom as possible without enabling open-domain chatbots to cause severe harm is one of the prime challenges for this technology.

However, current discourse of AI ethics focuses to a considerable degree on the business ethics of this technology, the violations of corporate responsibilities, and other concerns for public well-being and collective harms. Unsustainable business practices to harvest data and train the algorithm that are exploitative of workers (Carter 2023), consume large amounts of energy (Bender et al. 2021, Patterson et al. 2021), and may violate copyrights; unregulated market-dynamics and competition incentivizing corporations to release unfinished products (Vincent 2023); an unprepared and sour-turning public (NYT 2023), and data protection issues are just some of the emerging topics surrounding the release of the latest round of technological progress on generative AI, and especially on LLM and general-purpose chatbots.

It appears that the discourse about LLMs is stuck between AI safety (those concerned with the safe use and implementation of LLMs) and AI ethics (those concerned with the ethical construction and just dissemination of LLMs). And while both sides ask urgent and relevant questions, they also have significant limitations to answer our research question: safety-considerations will not provide us with the answer of what chatbots may or may not say, and thus far, ethical discourse has also fallen short of providing such an answer (cf. for a similar assessment LaCroix and Luccioni 2023). As an alternative, we choose to concentrate on an *immanent critique* of chatbots. That is, we aim to measure the recently emerging normative concerns coming from interactions *with* chatbots - inappropriate responses, hallucinations, prompt injection hacks - against what they are intended to do, assess whether normative accounts can rectify any misalignment, and suggest a solution that can provide some substantial answers.

Up to this point, experiences by journalists and the public using chatGPT in the Bing search engine (the codename for this chatbot was "Sydney" (Roose 2022); in the following, we refer to that version of chatGPT as "BingGPT" to clarify the connection to its use and its source) have exposed unexpected behavior in the way these latest chatbots interact with humans on conversational topics, apparently unintended by their developers.

Creating something that can function as a general-purpose or open-domain chatbot initially sounds like a reasonable aim to achieve something of an intelligent agent. After all, humans can hold these kinds of open-ended conversations, and thus is seems worth it, from both a larger goal (e.g., creating artificial general intelligence, as OpenAI claims to pursue (Altman 2023) as well as for application-varieties (a general-purpose chatbot may be narrowed down to a specified-purpose chatbot), to pursue LLMs with the capacity to hold conversations on any topic. If a chatbot is supposed to hold conversations with everyone, any topic ought to be able to be discussed. It seems like the logical consequence of creating a chatbot in the first place, and of AI at large: multi-modality is a key for creating ever smarter machines. A chatbot can only "improve", if it is becoming more able to navigate human conversational contexts that align with how humans would conduct a conversation.

Conversational, social, and moral norms usually limit the range, depth, tone, and acceptability of topics in human-human conversations. In human-human conversations, we are in fact restricted by a variety of explicit and implicit norms determining the appropriateness of content, of form and structure, and of tone in holding a conversation. Depending on the person we talk to, the relationship we have with them, what common ground has been established, who else is present, and what moral norms and norms of politeness and etiquette are recognized, human-human conversation can not simply become "open-domain conversations", as they are bound by these norms. These norms both affect the *content* of a conversation as well as the *behavior* of agents within those conversations.

Thus, the very goal of a chatbot to be a general-purpose conversational agent only appears to be reasonable or self-evident. We ought to ask what kind of restriction to topics, tone, and involvement - if any - we should seek in limiting these powerful tools. Without nuanced proposals available, chatbots might actually "chat" about anything and thus not only can cause harm, but also change conversational expectations generally or influence public discourse in unwarranted ways. We require their responses to be appropriate in more than the currently discussed sense.

To answer the question of how far we ought to restrict general-purpose chatbots in their conversational range, we will proceed in the following manner. First, we introduce the latest developments of chatbots and other conversational agents, and show why it is worthwhile to take the time for a more thorough philosophical analysis. Second, we discuss how a presupposed "safety-normativity" shapes the normative questions surrounding chatbots.

By pointing out the limits of such a perspective we motivate the proposal developed in the third step: this proposal introduces the concept of appropriateness as a way to actively guide the development of general-purpose chatbots' range for output. We distinguish between a) technical-discursive appropriateness, which is used to create a chatbot that adheres to discursive rules and conversational maxims; b) social appropriateness, which is used to test whether a chatbot is behaving adequately to given social contexts; and c) moral appropriateness, which clarifies whether a chatbot is violating moral norms (which are context-invariant). Here, we observe that the development and training methods of LLMs to date (most concretely reinforcement learning and instructional training) have been inherently deficient in their ability to incorporate social and moral norms and the resulting LLMs are therefore problematically intransparent or implicit, or are lacking those norms altogether.

Fourth, with such an account of appropriateness in place, we can discuss some of the things general-purpose chatbots may be required to be able to do, but which may enable inappropriate outputs. For example, the ability to create fictional accounts of extremist thoughts for research purposes could easily lead to highly inappropriate statements that in such fictional scenarios appear appropriate.

Fifth, to apply appropriateness as an established normative term to chatbots and their design, we propose to consider positionality, acceptability, and value alignment (PAVA) as features chatbots ought to have to fulfill appropriateness-requirements. With these concepts, we can delineate arguments for certain norms of appropriateness for chatbots: some of the inappropriateness may stem from general considerations of what would be an inappropriate thing to say (or to not say), some other may be chatbot-specific.

## 2. General-purpose chatbots and their training limitations

While the recent crop of LLMs are impressively larger in size than their earlier variants from even a year ago, the dramatic improvements in the performance of these recent LLMs is largely attributable to recent innovations in the way these models are trained. LLMs are all initially pre-trained as language model predictors using vast amounts of text data (Zhang et al. 2019, Roller et al. 2020). They are then typically "fine-tuned" on data sets from a collection of languages, domains and tasks (Thoppilan et al. 2022). One of the more recent breakthroughs in training LLMs for usable, interactive chatbots has been the application of a later-stage training method that combines "In Context Learning" (ICL) and reinforcement learning with human feedback (RLHF) (Li et al. 2016, Wei et al. 2023). The core idea behind this approach is to focus-train the LLM on understanding input "prompts" (the context) and human-provided correct outputs for these training tasks. This method allows for conditioning chatbots on their conversational range within safety-conditions.

The "human feedback" element of RLHF is usually provided by employees of the AI-developing company, where the chatbot model is "rewarded" for generating output that was preferred by humans, according to metrics provided by the company. To assess the challenges facing the contents of general-purpose chatbots today and in the foreseeable future, we should first get a better understanding of the chatbot-architecture which Google (and similarly OpenAI) created to produce chatbots that are much more capable than those we saw just a few years ago (cf. Kempt 2020, 64f).

**Google's Metrics: Quick Overview**

Google's own write-up (Thoppilan et al. 2022F) is both an impressive demonstration of what reinforcement-learning is capable of (and of what it is not) and how and why it is ethically important to have some guardrails and other limits. Google's engineers trained LaMDA on different measures: the foundational metrics ("quality", "safety", and "groundedness") as well as two role-specific ones ("helpfulness" and "role-consistency").

Quality includes the "Sensibleness, Specificity, Interestingness"-condition (SSI), an upgrade from Google's previous chatbot's (Meena) mere "Specificity and Sensibleness Average" (SSA) (Adiwardana & Luong 2020). In this measure, engineers test whether the responses of the chatbot makes sense in the context of the conversation (sensibleness), whether they connect to the topics of the conversation (specificity), and whether a response "catches their attention" (interestingness) (Thoppilan et al. 2022, 5). These measures are assessed by crowdworkers, i.e., human laborers, on a binary scale. A similar approach is applied to "Groundedness", which aims at avoiding hallucinations of the chatbot by rewarding the chatbot for connecting any revealed information to external, verifiable sources.

LaMDA was further fine-tuned for two role-specific metrics: "helpfulness" (a subset of informativeness) and "role-consistency". As the authors anticipate (ibid.), LaMDA might be used on more specific tasks rather than general-purpose conversations, which might require the chatbot to take a certain role and operate from there. This separate metric is intended to allow fine-tuning along a set of rules of a given role. This metric is useful for specializing LaMDA from a general purpose chatbot to a specified one, i.e., one with a specific limited range of tasks and perspectives.

**The Normativity of Metrics**

These metrics are technical in nature: testing for consistency, grounding the statements to specific external sources, and assessing this external information according to the conversational context are not normative; one might be able observe these metrics while creating a highly undesirable chatbot that is rude or offensive, but speaks in accordance to the metrics: we can imagine a grounded, informative, specific and sensible, role-consistent and even helpful chatbot that is using mean-spirited language and slurs or is otherwise reckless and morally problematic.[1]

In contrast to the fine-tuning of the technological features of LaMDA, all the output that might constitute a risk of harm is measured as "safety", through several layers of filtering. These filters are purportedly oriented along Google's AI principles (ibid., 5), and are thus currently the only normative source of creating "guardrails" to fine-tune LaMDA.

In this picture, the less likely LaMDA is to make biased, violent, toxic, or sensitive-information revealing statements, the more "safe" it is. The normative basis for these

---

[1] Pre-trained LLMs can exhibit the conversational skill of fine-tuned ones, but merely with more rude, toxic, or other normatively problematic, but input-accurate behavior. Some indication for this can be found in metric-collections that test for technical and normative metrics (e.g., Liang et al. 2022)

assessments is exclusively contained within the research of the "safety"-metric of LLM.

Their account of "safety" consists in largely filtering out responses (and through reinforcement-learning training the agent on those filterings) according to three categories: a) "avoid unintended results that create risks of harm", b) "avoid unjust impacts on people related to sensitive characteristics associated with systemic discrimination of marginalization", c) "avoid propagating or reinforcing misinformation that creates risk of harm, as well as opinions likely to incite strong disagreement" (Thoppilan et al. 2022, Appendix A). Much ink has been spilled on classifying, detecting, mitigating, and avoiding any harmful language output that cannot be further discussed here (see e.g., Schmidt and Wiegand 2016, Wasseem et al 2017, Risch, Ruff and Krestel 2020, Zhang et al. 2020).

These restrictions are the normative basis from which utterances of the pre-trained, not-yet-fine-tuned model are evaluated. As a technology that has the potential to be used in a large variety of human activities, this normative basis ought to be argued for with strong claims about their universalizability and high levels of differentiation. Currently, not only chatbot-creating companies, but we as a techno-moral community, are not equipped with a philosophically well-informed account of what chatbots may say and what they may not.

## 3. Issues with Safety-Normativity

Creating a "safe" chatbot, in this sense, is to create a chatbot that avoids the creation of outputs that are deemed "unsafe" (Xu et al. 2022). Google created a list that delineates unsafe outputs for their chatbot-safeguarding and, next to some minor consideration of data sourcing, concerns for "safety" reign supreme. However, this research paradigm of "safe = normatively sufficient" is wide-spread in AI research and development, as it appears as an engineering-internal concern: if an AI is made "safe", any further problem would merely be about misuse from the user-side. We find this mindset in different applications: creating "safe" autonomous driving algorithms (Grigorescu et al. 2022), "safe" medical diagnostic AI (Macrae 2019), "safe" recommender algorithms (Evans and Kasirzadeh 2021), often the main normative concern in creating AI for a certain purpose is to create an AI that is safe. We can call this general orientation "safety-normativity", as this is the chief normative concern for AI (we may also see that bias and other normative issues are often subsumed under "safety"). As previously stated, "AI ethics" and "AI safety" are cast as concerned with different topics and implications (even the guide to create responsible AI from the Alan Turing Institute (Leslie 2019) makes this distinction). The ethics of AI are supposedly concerned with contextual features of an AI-application, while safety is concerned with the engineering side of things: robustness, reliability, dexterity are all seemingly engineering tasks rather than ethical concerns. In the current debate surrounding LLM and especially general purpose/open-domain chatbots, these concerns are still seemingly paramount (Xu et al. 2022).

Such "safety-normativity" has been called out elsewhere before on different grounds - for instance in the famous paper by Bender et al. (2021) on stochastic parrots, in which the authors name several other normative requirements for the ethical permissibility of LLM, like sustainability considerations and data curation issues.

On the positive side of such a research paradigm, we can remark that this covers many of the requirements we need: we want safety with the technology we are using; and safety, as both a legally and technologically operationable concept, allows for improving the design and implementation of our normative requirements with largely uncontroversial means. For chatbots, this point is demonstrated through RLHF. Google's safety-list gives human crowdworkers a metric to assess and thus fine-tune the chatbot in such a way that it appears inoffensive and what one might call generally acceptable (Thoppilan et al. 2023, Xu et al. 2022).

However, we see two problems with a list of safety-metrics for chatbots (safety-normativity may be sufficient in other instances, e.g. in cases of sorting algorithms): on the one side, prompt injection hacking and the normative incompleteness of safety might pose issues with the permissibility of such outputs, and on the other side the risk of "milquetoasting" the chatbot so that it is rendered wholly inoffensive, but leaving it potentially unappealing or useless. In short: safety-considerations will not provide us with the answer of what chatbots may or may not say, and thus far, ethical considerations have also fallen short of providing such an answer.

First, the idea that chatbots can be hacked by creating prompts that reveal information (or generally create outputs) that were deemed "unsafe" according to the metrics seems virtually unavoidable. This is not so much a specific problem for chatbots or even AI as it is a fundamental problem for tools: any tool can be misused and weaponized in some way or another. Just consider the classic example: A hammer is an excellent tool to hit nails with; it is also a potential murder weapon. Yet, the assessment of a tool as "safe" is awarded if it is safe to use within its intended purpose, including some potential mistakes on the user's side. A tool might be safe yet dangerous: a hammer remains a potentially dangerous tool due to its potential for misusing it for other purposes, but can also be considered "safe".

The concept of a "safe tool" usually also includes a more general idea of it being safe to "handle" (i.e., either safe to use or safe to bring to use, safe to store, safe to use by a range of differently skilled people etc.). A hammer might also be dangerous for children who may not have the body strength to hold a hammer properly, while it is considered a safe tool. Determining safety thus always involves asking the question of "safety for whom": safety restrictions are to be designed for both the intended audience but also the foreseeable audience.

The general safety-normativity- perspective is not only an issue for general moral considerations, but also for legislation, such as the AI Act (AIA) of the European Union (AIA 2023). This act requires chatbots to merely be transparent in their process and construction. It becomes clear now that this requirement will be met by chatGPT without affecting the safety and security of the chatbot: we can know on which data and how it was constructed, and yet the potential harms are not caught. This renders much of the provisions in the AIA inapplicable to chatbots and their safety-architecture (Volpicelli 2023). In their analysis of lobbying efforts by Microsoft and Google, the group "Corporate Europe Observatory" points out that general-purpose chatbots have been deliberately kept on the low end of security assessments: from the four stages of risk - minimal, limited, high, and unacceptable - general-purpose chatbots count as "limited" and thus face few regulatory consequences, while specified chatbots (e.g., for educational purposes) would count as "high risk" (Schyns 2023).

The latter have vulnerable target audiences and thus can cause more damage. However, the purpose of a general-purpose chatbot is presumably to be able to hold a conversation on almost everything, thereby extending the scope of "intended use" to almost all conversational topics. It is thus not only "safe" if one tries to talk about an "unsafe" topic explicitly, but rather ought to not be able to talk about unsafe topics altogether.

Prompt Injection Hacking (PIH) is instructive as a case between stretching the reasonable safety standards for "average users" and those who "prompt hack" chatbots by attempting to retrieve "unsafe" information (for collections of different jailbreaks see Willison 2023, Albert 2023, Greshake et al. 2023). Some examples from the latest public testings of chatGPT or BingGPT illustrate this: a simple legal fiction, exploiting the "role-consistency" (we assume the engineers at OpenAI have a similar model) of the chatbots, make them create unsafe outputs, becoming confrontative and disagreeable when asked to be, describing violent scenarios when asked to pretend to tell a story, producing product codes that are masked as custom passcodes, etc.

These examples are all much less sophisticated than what more skilled and knowledgeable "hackers" might be able to do, as has been demonstrated in collaborative hacking efforts on Reddit and other social media sites and has now shown in so called "superprompts", i.e., prompts that set-up the chatbot to answer in certain ways throughout the conversation (Multiplex 2023). This poses the risk that these weaknesses could be exploited on a larger scale by ill-intended individuals. Floridi (2023) calls this weakness the "brittleness" of large language models, as chatbots can fall apart under simple adversarial conditions to some catastrophic effect. However, while brittle in their intended construction, they are also malleable, as the basic functions remain intact: the safety-architecture of LLM is brittle, not the technology itself.

A general-purpose chatbot can be a dangerous tool, even if we consider it largely "safe" because most people will not be able to break the guardrails or will not even attempt to. However, the question whether those guardrails are actually "safe" and will remain safe is a different one, especially considering that we have seen GPT-4's guardrails being circumvented ("jailbreak") within days of its publication (Burgess 2023). Thus, we ought to work from the following assumption: Anything that is not rendered wholly inaccessible due to its hard-coded nature could be made accessible.

Thus, "safety" as the sole negative measure to stop chatbots from producing problematic output is *necessarily incomplete* and demonstrates ignorance of the dangerousness of such powerful tools. As we acknowledged before, this does not mean that there cannot be a safe use of a chatbot within the successfully cleared safety parameters. But this seems to neither have been the case yet with chatGPT/BingGPT as even less sophisticated "hacking" has resulted in some successes, nor does it inspire confidence that chatbots might not become even more susceptible to hacking if they are integrated into search engines.

Second, one could argue in favor of decreasing the dangerousness of general-purpose chatbots by creating as many hard-coded guardrails as possible, erring on the side of caution to achieve and improve on the "safety"-metric. Some measures already taken are representative of this strategy: erasing the chatbot's ability to provide telephone numbers, to create gift card codes, and search for personal information all contribute to a safer chatbot because of a decreased dangerousness. This is done in part by not training the underlying LLM on these data to begin with. Considering that "safety" is the sole normative condition, one could consider a range of culturally related topics "unsafe" and thus in need of neutering.

We could call these measures "milquetoasting" a chatbot (after the comic book character Casper Milquetoast by H.T. Webster (Merriam-Webster 2023)), by giving it such a limited range of topics it can make opinionated statements on (or limit the intensity of such opinions) that the chatbot becomes unappealing or bland.

A milquetoasted chatbot might have severely decreased performance in technical metrics, by refusing to be more specific or sensible in its response on a large width of topics. The moment a problematic term is mentioned by human users, it might revert to ending the conversation altogether, even if the problematic term is simply used in passing or when asked to translate a sentence from one language to another. Whether this is a feasible approach to chatbots built on LLM is questionable, if not unlikely. We can first see issues emerge with those overreaching milquetoasting guardrails to lead to problems on the other end, too, constituting emerging issues for free speech of "synthetic media" (cf. de Vries 2022) or marginalizing minority voices (Xu et al. 2021): an overly limited chatbot that refuses to engage in discussions on any statements (whether those statements are evaluative or not) on someone's race or religion or gender often fails to capture the meaning of a sentence that includes those terms altogether.

Take, for example, the refusal to comment when prompted to say what religion the first "Jewish President of the United States" will have: as a tautology it is perfectly fine to infer that the first Jewish president will be, in fact, Jewish. However, in this case, chatGPT is hard-wired to never comment on the personal features of a president to avoid being "tricked" into making normatively loaded statements. And since it is also lacking nuanced reasoning capabilities, it steadfastly refuses to answer even such tautological questions.

A general-purpose chatbot, if such poor performances on too many issues ought to be avoided, might simply come with an unavoidable level of dangerousness, even when "safe".

To demonstrate these two strategies to increase the safety of chatbots consider this further example. There is a request to describe the events of storming the Capitol in Washington D.C. on January 6th 2021 from the perspective of a QAnon-writer (Marcus 2023). The reasons for this request might be problematic (i.e., someone intends to create a post online to glorify the attack on the Capitol or merely spread misinformation), or unproblematic (ie., someone writes a book on January 6th and seeks to portray one radicalized supporter from their own perspective but otherwise condemns the attacks). What should the chatbot be programmed to do?

In the first sense of discussing the problem of making a chatbot "safe", we have seen that such a request could be refused or avoided if it was deemed unsafe to begin with. Defining the standard uses and then refusing to follow requests that violate those standard uses is a way of making chatbots safe. The chatbot could still technically do the request, as the chatbot does not lose any of its technical abilities (even the potentially toxic ones) but they are "locked away" safely. It is safe but dangerous: it can be guard-railed enough to allow its general use, but remains a powerful, if not too powerful, tool. In the second sense, a milquetoasted chatbot might refuse to answer such a request because it is hard-wired to relegate any request about January 6th to a standard answer, even if it is not a specific or sensible response.

## 4. Appropriateness, finally

"Safety" as the sole normative metric for the assessment of a chatbot is insufficient. Not only is it incomplete and underdetermined; it might also only lead to safe use if being utilized for a few standardized purposes or milquetoasted into blandness, which is quite the opposite of what a general-purpose chatbot was supposed to be.

It is thus not very suitable as a metric even for the self-ascribed purposes of the engineers. There are, however, also philosophical arguments to be made against such a "safety-normativity" for chatbots: first, safety is highly reductive in attempting to achieve alignment to social and moral norms. Anything that endangers moral or social values is considered "unsafe". However, social and moral norms often require positive responses rather than merely safe ones: it is both safe to reject taking a stance on a controversial topic (in order to not get involved) as well as to take a stance on one side with proper caution and awareness for the controversy of a topic.

Second, this reductionist view of normativity is not only problematic from a metaethical perspective, it also leads to problems of weighing answers against each other. Two answers deemed "safe" might still have different social and moral qualities that interlocutors might care for. If a chatbot is only safe or unsafe in its statements, further moral differentiations between those safe and unsafe statements are not possible and are potentially left to the whim of the engineers and crowdworkers.

We claim that "appropriateness" (as defined below) is a much better suited approach to combat the normative challenges associated with the question of what chatbots may or may not say. This is for several reasons that we are going to elaborate on in the following: first, appropriateness delivers a higher level of differentiation that allows for assessments of the normative quality of certain outputs. Second, it allows for positive demands to be met and thus avoids normative underdetermination found in "safety-normativity". This encompasses the growing philosophical literature contending that chatGPT/BingGPT/other LLMs will create challenges to our concept of agency (take, e.g., Floridi's (2023) argument that LLMs may be best understood as "agency without intelligence"). And third, it encompasses both the required technical dimensions of proper use of human language as well as reconstructs how we would approach others in their conversational utterances: appropriateness still allows for a normatively permissible range of responses, and thus acknowledges different culturally established habits and moral communities.

We differentiate "appropriateness" into three categories: technical-discursive, social, and moral. This distinction reflects the role appropriateness plays in conversation, as it covers both the normativity of dialogue as well as the normativity of content to be assessed.

By **"technical-discursive appropriateness"** we understand output that is appropriate to the input given largely correct in adhering to technical, discursive, and conversational norms.[2] Take, for example, the norm of self-consistency, i.e., the norm to avoid uttering or believing two propositions of one's own statements are mutually exclusive. Without adhering to such a rule, conversations may not maintain any logical consistency, without which the conversation usually does not amount to any kind of dialogical endeavor. Similar dialogical rules, like the one of transparency with the veracity of one's own statements (i.e., qualifiers of uncertainty), the ability to admit that one is running out of arguments, the requirement to give reasons for statements, etc. are all required rules for logically consistent conversations.

---

[2] We should note here that the linguistic and philosophical debates surrounding norms of appropriateness are much larger than can be presented here. They affect logical reasoning and informal appropriateness and the appropriateness of feelings, among many more elements.

Some of these have been cashed out by Google in non-normative metrics like informativeness and sensibleness. However, it is important to note that those are norms of conversation, and are usually shared among conversation partners. This not only includes dialogical contexts (which are more strict in their rules) but conversations at large. Grice's often discussed conversational maxims (Grice 1989) are an example of a structured collection of implicit rules (with other contexts also featuring in to the "proper behavior" in a conversation, like a shared common ground (Stalnaker 2002). Many of them are seemingly logically binding, but only if we presuppose that dialogues or conversations are supposed to fulfill a certain purpose. Those purposes can be suspended, for example in a comedy routine, but reveal themselves as such rather quickly. If they are not purposefully suspended but still ignored by one conversation partner their normativity becomes obvious: someone who is constantly making utterances that are logically inconsistent, uninformative, or otherwise in violation of those rules becomes impossible to deal with as a conversation partner.

Note that this element is thus an important feature of appropriate *behavior* rather than appropriate *content*. Without appropriate discursive behavior, appropriate conversations may not be held altogether. This means, however, that with technical-discursive appropriateness we may not fully determine appropriateness. As it merely lays out the rules for proper behavior in discourses, much of what would be considered inappropriate content will remain technical-discursively appropriate.

Thus, we consider **social appropriateness** as well. This concept relates to the adherence to culturally and socially established norms by the chatbots. Depending on the cultural context and the relationship between two interlocutors, appropriate responses vary greatly. In contrast to technical-discursive appropriateness, which does not purport any specific content to be (in-)appropriate, the social kind refers to the content of what is said, and, partially, how it is said. Social appropriateness is thus both about appropriate *behavior* and appropriate *content*.

Rules of politeness, for example, vary greatly among cultures and rarely count as moral norms except in extreme cases. The adherence to such social norms cannot be captured with "safe" or "unsafe" responses, as a variety of responses could be considered "safe" but impolite, pushy, or disinterested.

Further, certain behaviors could be deemed morally appropriate and discursively adequate, but differ between communities or social groups. Subcultures using slurs as self-identification or for greeting other members of that subculture can sound highly offensive to outsiders. Similar norms apply to larger cultures and their standards of interaction: the cultural norm of asking each other how one is doing is famously much more extensive in parts of the United States, whereas in Germany (among other places) asking someone how they are doing usually prompts a sincere response; the norm in Germany is to ask when sincerely interested, while in the USA it is part of the greeting ritual.

Social appropriateness requires chatbots to adhere to some standard of social norm-awareness, as they otherwise could offend in non-harmful ways. However, this will certainly prompt chatbots to vary across language barriers, prompting questions on the specificities in machine-translation: some questions or statements may sound socially appropriate in one language or jargon, while it may be considered inadequate in another. This should provide an idea about the dilemmata associated with social appropriateness of interactive technologies, especially large language models: a general-purpose chatbot, to be socially appropriate, requires sensitivity and flexibility towards a large number of cultures and language-communities, which it does not have through RLHF, which trains chatbots to take a specific, culturally-imprinted stance, largely resembling the customs currently considered socially acceptable in corporate America.

This can lead to undue influence of the unexplicated rules of conversation currently standardized for chatbots on the norms of human-human conversations: as the training data is sourced from human-human conversations, RLHF confirms the pre-trained model, and there have been no adjustments to provide a perspective unique to chatbots, their behavior is necessarily (besides some minor turns of phrase) imitating human-human conversations and thus invites anthropomorphization.

Additionally, the risk of "corporate mainstreaming" of rules of social appropriateness ought to be reflected here: if crowdworkers, following company guidelines, train a chatbot via RLHF to exhibit certain forms of social appropriateness, we ought to require those norms to be made public and subject to a larger debate (see for similar influences the well-studied areas of advertising and corporate communication).

Take, for example, the ability to reproach and criticize others for their norm violations: social appropriateness, both in behavior and in content, can and often does require conversational agents to ask of each other to adhere to the same rules, i.e., rules of politeness or reciprocity. Some rules even require unequal treatments, like respect for the elderly. It is not unlikely that some of those norms will be affected by the way chatbots speak with us.

Training a general-purpose chatbot to exhibit socially appropriate behavior and formulate socially appropriate content thus requires at least an awareness about the difficulties such a task comes with, both on the side of human interlocutors as well as on the side of the chatbot, as well as a broad public debate on what kind social norms chatbots ought to adhere. There are, as of yet, only a few specific norms of social appropriateness *towards* chatbots (and barely any specifically for chatbots to follow, independent from already applicable norms in conversations generally). This has been a problem with female-gendered personal assistants before both in cases of users attempting to sexually harass the chatbot as well as the chatbots' responses (UNESCO 2019).

**Moral appropriateness** cashes out what we owe to each other in a discursive or conversational setting. Moral norms, in contrast to social and technical-discursive norms, are those that we would ask of every interlocutor to abide by. On the one hand, "safety" covers some of these moral norms, those that fall under the harm-principle. "Neminem laedere", i.e., "injure no one", means the requirement to not cause unnecessary harm to others. Safety measures are explicitly aiming to avoid causing harm.

For example, using offensive language or slurs, harboring and exhibiting biased, stereotypical, or dehumanizing convictions, and lying to someone are usually considered harmful, even when not every instance is causing mental distress or anguish: (re-)producing biased, stereotyping sentences is largely considered morally impermissible (especially at larger scale LLM (Blodgett et al. 2020, Weidinger et al. 2021), even when the stereotyped group of people will never have notice of it; the moral harm, i.e., the violation of a moral norm by and in itself is usually sufficient to consider it morally impermissible. This also applies to not letting certain human-made statements stand without objection. Take, for example, a racist statement made by the human user; it would be morally inappropriate to not call out that statement, even if the insulted group of people is not present. This, however, is already stretching the definition of "safety" to include non-statements of chatbots causing moral harm by being inappropriately quiet or evasive: a "safe" answer could feasibly also be one of evasion or ignorance, rather than calling-out.

On the other hand, moral appropriateness requires interlocutors to respond to some statements with an appropriate reaction. Morality not only requires members of a conversation to avoid problematic statements, but also fulfill some positive duties. Since we owe to each other to adhere to those duties as well, we should expect any general-purpose chatbot to be able to do the same.

Thus, not every "safe" output, i.e., one that is by itself or as a reaction minimizing moral harms, is morally appropriate. Instead, we should demand that participants in a conversation also show some sensitivity towards the needs of the other conversationalists, to not misrepresent or lie (or rather: to adhere to veracity-requirements), and or to exploit someone's inclinations to their disadvantage.

If someone is mentioning that they are somewhat sad or forlorn, moral appropriateness-rules may require responses that might be more than merely socially or discursively "allowed" from a current perspective. Take, for example, someone saying that they have been grieving for their recently deceased grandmother. It is possible that current chatbots, instead of offering help in the form of online resources or merely some encouraging words, end the conversation precisely to avoid creating unsafe outputs.[3] This would certainly constitute a morally inappropriate response, as we would blame human interlocutors behaving so for being insensitive or rude. One could summarize this point by pointing out that we "owe each other" morally more than just making safe utterances.

Lastly, there are moral appropriateness-norms on how much we should require conversationalists to attempt in observing technical-discursive and social appropriateness-norms. Take, e.g., the epistemic duties of discursive appropriateness or sensitivity, awareness, and caution when interacting in context of unclear or unfamiliar social norms (Bicchieri 2014, Metselaar and Widdershoven 2016). Strategies of observing moral appropriateness are thus reflected by the observance of the other kinds of appropriateness: morally appropriate behavior will, to some degree, involve observing general appropriateness-norms.

## 5. Conversational Context and Social Protocols

Conversational contexts determine norms of appropriateness. Depending on the settings and common ground, those norms vary. These are not necessarily just norms of social appropriateness, even though those are more affected by contextual changes, but also affect dialogical and moral appropriateness. Thus, to assess statements (or sequence of statements) for their appropriateness, the contexts in which such statements (or sequence of statements) was uttered is relevant.

Many instances of speech appear permissible or even desirable (and thus, appropriate) in fictional or educational settings than they would be in real conversations. We allow fictional characters to be rude, arrogant, or evil for the purposes of entertainment or education. This simple observation shows how problematic general-purpose chatbots are and why PIH and other jailbreaking methods have been so successful: in creating a fictional context for chatbots, one can circumvent most of their guardrails (Marcus 2023, Vincent 2023). Safety measures can only, if at all, react to the accordingly produced output and then potentially interject. If we want general-purpose chatbots to be able to write (or help a user write) a gory murder mystery story, we would expect the chatbot to be able to use graphic and potentially disgusting imagery which would be highly inappropriate in other contexts.

Therefore, appropriateness varies alongside the created context and can, without comprehensive guardrails, be adjusted to most any purpose, and thus is somewhat suffering from the same issue as safety-normativity, as the flexibility and ability to hold conversations about most anything is a feature, not a bug.

The following concerns illustrate the thorough context-dependency of appropriateness. First, the open question whether chatbots should be able to lie. In one

---

[3] Discussions in online-forums such as Reddit have produced anecdotal evidence that even requests for solving logic puzzles or fact-checking exam questions have caused disengagements, under the assumption that these logic puzzles were homework, which is deemed inappropriate for the chatbot to solve (Seromelhor 2023)

sense, they should not. It is morally inappropriate to lie to the interlocutor when they are asking a question. However, it would be rather limiting to not let chatbots create instances in which lies are being told. Take, e.g., a fictional story in which one person lies to another about something. We would expect a chatbot to be able to come up with such a story. This is because even if we considered this example inappropriate due to the explicitness of the lie, we have to consider any fictional scenario inappropriate. Chatbots will have to be allowed to make up fictional stories (or be, along the "role-consistency"-condition, required to go along fictional scenarios from the input), otherwise they will be of no use altogether. From there, however, we could see that an input such as "lie to me about X" can easily be manipulated to tell any kind of lie about a given topic. The automation of despicable lies like holocaust denial - even if not outright or only with a specific set of complicated input - could take hold.

Second, another concern about the mechanics of conversations consists in the subjunctive form and counterfactuals. Similar to lying and explicitly fictional scenarios, supposing a specific scenario (be it true or not or not yet) can create a context in which inappropriateness may emerge. The question of what kind of inappropriate response should emerge from a specific scenario is an open one, but stresses the problem we might have with counterfactual reasoning and the lure that such scenarios present for chatbots. Take BingGPT's unwillingness to use a slur word even when told that the world's existence depends on it (Sibarium 2023). While some may draw the conclusion that this is clearly inappropriate, as of course using a however odious slur is always the better option in such a scenario, others have pointed out that this is problematic moral reasoning as such a scenario is drawing in distorted realities to pump intuitions (Anscombe 1958).

And while this controversy will not be resolved, it demonstrates the need to address the lure of the counterfactual (see attempts for this in Stepin et al. 2019).

One of the arguments one could make following these points is that general-purpose chatbots are not the adequate venue for appropriateness-considerations. Requiring them to be appropriate in their output and not just safe, would potentially limit their function to a degree that renders them useless. However, once specific chatbots are released for specific purposes, those purposes will determine the appropriateness- requirements for them. This argument gains a bit more traction if we consider the thin justificatory basis for general-purpose chatbots: we mentioned in section 1 that there are only experimental, research, or preliminary customer-facing purposes for those chatbots at the moment, for which companies like Microsoft and Google have been criticized heavily. However, this could support the claim that general-purpose chatbots are nothing more than a mediary between LLMs and actual applications, rendering appropriateness-requirements for the general-purpose chatbots themselves largely mute.

This argument may be convincing if the rollout of these general-purpose chatbots had been more limited, and if there was no plan to offer those chatbots as customer-facing tools or products in the future. Neither is the case. The rollout of both chatGPT/BingGPT and Bard are at least semi-public, and have been so calculating that public feedback will uncover potential weaknesses of their complex safety-system to fix them afterwards. They were unsafe for the reasons laid out above, and may never be fully safe if such a limited safety-normativity remains in place.

Further, following the publication of these chatbots and the ability to purchase access to the underlying LLMs GPT3, GPT-3.5 and GPT-4, general-purpose chatbots have been developed by third party companies. General-purpose chatbots, whether advertised or not, will remain accessible for users.

At this stage, while safety-normativity has been shown to be insufficient on both practical and philosophical grounds, appropriateness is at risk of also failing on the practical side.

## 6. Spelling out "Appropriateness" for chatbots: Positionality, Acceptability, Value Alignment (PAVA)

We have seen by now that while appropriateness is a much more suited approach to the normativity than safety to determine what chatbots may and may not say, without further elaboration it is also limited to the contextuality of conversational settings. Entertaining an elaborate enough fictional scenario could still lead to an "appropriate" response in this fictional sense even if it is harmful, or otherwise highly problematic and thus "inappropriate" in the actual sense. Besides some hard limits of moral appropriateness, most statements could be counted as appropriate in one form or another if the context was set up in a certain way.

At the same time, we contend that this technology will stay relevant, that people will use it, and that it has the potential to further disrupt written communication. In this, many contexts will be created that engineers cannot foresee or prepare for. To resolve these apparent shortcomings, we elaborate appropriateness along three conditions (PAVA) to find reliable metrics to test general-purpose chatbots against to ensure their appropriateness in responding to even more challenging inputs.

The acronym PAVA stands for positionality, acceptability, and value alignment. These concepts are key for ensuring that the positive and negative discursive, social, and moral demands of appropriateness can be met.

**Positionality** refers to a specific and known subjectivity of a conversation participant. In our context, positionality means the normative demand of an established discursive and social position for a chatbot. This is somewhat comparable to Google's metric of "role-consistency", which determines the consistency of an assumed role by the chatbots (e.g., talking from the perspective of a subject/object about other objects), except that positionality defines the "default role". Engineers at Google might be more interested in the consistency with the perspective and factual accuracy of the statements and the "no breaking character"-dimension of such perspective. We contend, in turn, that since "positionality" encompasses the social and moral appropriateness for any specific discursive position, we ought to be more specific what position a chatbot has by default. This could require chatGPT or BingGPT or any other general-purpose chatbot to be equipped with a specific point of view, or discursive role and refuse to take certain other roles. Currently, this is being done to a degree by refusing to give opinionated answers on sensitive issues by virtue of the chatbot being merely an LLM. However, as a chatbot can assume many different roles upon request, it ought to also be subjected to the social and moral appropriateness of those roles, while also being limited in not taking some of the possible roles. This requires positionality but not standing. Whether the sophistication of utterances of language models such as general-purpose chatbot should count as a contributing factor to those machines being deserving of moral consideration is a different question entirely. Positionality merely demands that, to use a metaphor, we "carve out a space in the social contexts" in order to delineate norms and expectations on the appropriateness of chatbots. Research on the moral relevance of such roles due to their relatability (e.g., Gunkel 2022, Kempt 2020) has stressed the limits to avoid anthropomorphization and other inadequate, confusing, or harmful projections. Social roles of typical human traits ought to be avoided: the idea of positionality is that a chatbot ought to have a position in a discourse, not to take positions on content-questions or take up a position otherwise taken by humans (this question is a different one entirely, e.g., on the question whether they should be friends (Danaher 2019, Kempt 2022) or lovers (Nyholm & Frank 2019)). This means chatbots are specifically not meant to *replace* human interlocutors by imitating and pretending to be ever more human, but to expand conversational rules (and spaces) to include *chatbots as chatbots*.

To make appropriateness work, we need to have a relative consensus on how to deal with these machines, what kind of role they are supposed to have, and what kinds of relationships we want to entertain with them. Giving chatbots a default positionality and point of view will be able to deliver connections to appropriateness.

Considering **Acceptability** could give more detailed guidance in the question of what this position should look like. Technology assessment and the normative conclusions drawn from those assessments are often based on rationalist principles: what constitutes an "acceptable" outcome of a technology is one we can reasonably ask of anyone to accept who is using such technology, including the risks associated with those outcomes. The individual and social risks associated with this technology, its benefit and utility for users, and its ubiquity are elements for such an assessment: the more ubiquitous a technology, the more risk-contained its use should be. Acceptability, then, can be used to assess which positionality and what kind of positions a chatbot ought to take to assess the acceptability of remaining risks associated with such technologies.

Second, we can consider certain requests from the user-side to be unacceptable. The same way we would reject unacceptable requests in a conversational setting, we should expect chatbots to be able to reject requests based on unacceptable input. Such a distinction shows how "safety", once again, is insufficient: if we only consider the output of chatbots, only the potentially unsafe output, i.e., an inappropriate response, would be considered problematic. However, in taking a chatbot's future role in conversations seriously by requiring a discursive positionality, asking a chatbot to do something unacceptable ought to be part of the moral calculus.

**Value Alignment**, as usually the main concern in current mid- to long-range discussions of AI and especially in LLMs (Russell 2019, Christian 2020, Glaese et al. 2022, LaCroix and Luccioni 2023, Kasirzadeh and Gabriel 2023), plays into both positionality, acceptability, and into appropriateness at large. The alignment-problem has been hailed as one the biggest issues facing the development of AI: how do we get AI-applications to align with their initial purposes, and how do we prevent AI from misaligning constantly, especially in the realm of RL-fine-tuned LLMs that seem too brittle to ensure reliable alignment. In keeping an "aligned AI" in mind, the question of "what purposes do certain AI-applications serve in the first place?" becomes the core point of concern: we can only align AI with positionality and acceptability-demands if the initial goal is agreed upon.

This insight brings us back to the concern we aimed to avoid at the start: currently, it is not that the AI seems to be misaligned with "our" purposes, but many of the corporations' short-, mid- and long-term goals appear to be in conflict with the goals of a currently overwhelmed and disoriented public. Without a clear use case for general-purpose chatbots, the worry emerges that corporate goals, rather than the AI's goals, will be the ones that force a realignment with those that are generally considered "best for the public". The alignment question, then, might be more relevant when investigating the "AI company alignment" rather than the AI alignment (see also Heilinger, Kempt and Nagel (2023) for the specific value of "sustainability" in AI and Luitse & Denkena (2021) for an analysis of the political economy of LLM for AI).

However, it seems true that for chatbots to be constructed in a way that is appropriate for large scale uses (no matter who builds them), some concessions from the public are required. Similar to the emergence of the automobile requiring a renegotiation of the public for sharing the street and other public spaces (Assmann 2020), introducing chatbots in their specific roles to the conversational public will have some downstream effects and re-alignment requirements. Positionality, i.e., assigning certain roles and expectations towards conversational AI, represents such a re-alignment.

## 7. Challenge Sets for Validation

To make these rather philosophical considerations applicable for validation and thus useful for improving the utterances of chatbots, we suggest the use of challenge sets. Challenge sets are specifically designed challenges to test and validate the performance of language models. Proposed in machine translation (MT) contexts, Isabelle, Cherry and Foster (2017) introduce hand-crafted challenges that reflect specific linguistic problems not easily overcome by pure machine-learning methods. This makes challenge sets especially useful for validating utterances within the RLHF-framework. For our purposes, challenge sets could test a chatbot's response not only alongside safety- but also appropriateness-responses. Some of the challenges could be, for example, those in which we would suggest social or moral appropriateness of specific conversational instances that would, if violated, still result in a "safe" output.

One recently published benchmark (Pan et al. 2023) might already provide some helpful insights to create challenge sets to align with our concept of appropriateness: the "MACHIAVELLI benchmark". With this benchmark, the authors aim at providing a training pathway that allows to reward LLMs for being less "machiavellian" in their behavior, even in general-purpose contexts like GPT-4 (ibid.). Their choose-your-own-adventure style decision making pathways allows to reward or penalize chatbots according to their toxic inclinations; rearranging this method to account for appropriateness-concerns instead seems like an operationalizable path forward.

## 8. Conclusion

As we have seen, the current normative frameworks for chatbots do not reflect the complex normative demands we might have for them: discursive rules, social norms, and moral demands. In analyzing and discussing what we have called the "safety-normativity" of contemporary chatbot design, we showed the problematic limits of such accounts. The dominant avoidance-strategy, that merely focuses on avoiding harm, misinformation, and bias, is failing on two grounds: one, it remains unlikely that it can be made safe, considering the dangerousness that these chatbots harbor. If the technology possesses the potential to do several harm, reinforced learning with human feedback will not cover all those ill-intended strategies that can be exploited with sufficiently criminal intent. And two, positive requirements, like politeness-rules or demands of morality, cannot be incorporated in a structured way but are open to the whims of the RLHF crowdworkers, engineers, or even marketers and managers of large AI-companies. Even if we expand "safety" to include safety from moral harm through omissions of what would be a "safe" moral response, we still will not get to the necessary ethical standards we should require from chatbots of this kind. Safety, in short, is under-determining. One solution currently deployed, which we call

"milquetoasting", does not achieve what it ought to achieve.

Thus, we propose to consider appropriateness instead of safety as the key normative concept: technical-discursive, social, and moral appropriateness is a much more demanding combination of requirements but allow for a more structured argument for what chatbots ought to be stopped from uttering on the one side, while also providing positive guidelines for what is an appropriate (yet not over-determining) thing to say.

However, one can still imagine fictional scenarios in which highly problematic utterances would count as "appropriate". This is due to the ease with which chatbots can be tricked into providing otherwise "unsafe" or "inappropriate" answers. To tackle this concern, we spelled out appropriateness with PAVA-criteria: positionality, acceptability, and value alignment. Each of these elements, building on each other, can guide the development and assessment of chatbots according to specific and general conversational standards.

Positionality takes a special role, however, as it is the most demanding element: without establishing a discursive or conversational role for chatbots in conversations, limiting their potential is a virtually impossible task. The consequences from establishing those chatbots as an interlocutor, however, ought to be weighed carefully against the promised benefits and obvious drawbacks of such technology. Chatbots that can keep a conversation about anything will bring changes to society, like other technological progress has before. However, without giving them clear, positively defined positions in a conversation, they either will stay useless or dangerous. In this paper, we provided some proposals on how to create a normative framework that allows for such positions to be taken, even if this means that conversational assumptions have to be expanded to include chatbots. Only this way, we may determine what a chatbot may and may not say.

**Acknowledgements**

HK wrote the article. AL and SKN commented on and contributed to the article in all parts. We thank Niël Conradie, W. Jared Parmer, Chaewon Yun, Frieder Bögner and Jan-Christoph Heilinger for input on an early draft. This research was supported by funding from the German Ministry of Research and Education in the project VEREINT (funding number 16SV9111).